\documentclass[letterpaper]{article} 
\usepackage{aaai2026}  
\usepackage{times}  
\usepackage{helvet}  
\usepackage{courier}  
\usepackage[hyphens]{url}  
\usepackage{graphicx} 
\usepackage{natbib}  
\usepackage{caption} 
\usepackage{algorithm}
\usepackage{algorithmic}

\usepackage{newfloat}
\usepackage{listings}
\DeclareCaptionStyle{ruled}{labelfont=normalfont,labelsep=colon,strut=off} 
\floatstyle{ruled}
\newfloat{listing}{tb}{lst}{}
\floatname{listing}{Listing}
\usepackage{multirow}
\usepackage{tabularx} 
\usepackage{amsmath} 
\usepackage{svg}

\usepackage{xcolor}
\usepackage{amssymb}
\usepackage{bm}
\usepackage{array}
\usepackage{color}
\usepackage{paralist}
\usepackage{booktabs} 
\usepackage{arydshln} 
\usepackage{tikz}
\usepackage{makecell}
\usepackage{colortbl}
\usepackage{amsfonts}

\usepackage[switch]{lineno}
\definecolor{red}{RGB}{255,44,0}
\definecolor{ired}{RGB}{229,72,72}
\definecolor{igreen}{RGB}{80,219,144}
\definecolor{BlueGreen}{RGB}{40,200,148}
\definecolor{NavyBlue}{RGB}{0,176,240}
\definecolor{lightblue}{rgb}{0.93, 0.95, 1.0} 
\definecolor{uclablue}{RGB}{203,234,250}

\newcommand{\circlenum}[1]{%
    \resizebox{!}{0.8em}{%
        \tikz[baseline=(char.base)]{
            \node[shape=circle, fill=black, inner sep=0.8pt, text=white] (char) {#1};
        }%
    }%
}

\title{Orthogonal Spatial-temporal Distributional Transfer for 4D Generation}
\author{
    Wei Liu\textsuperscript{\rm 1},
    Shengqiong Wu\textsuperscript{\rm 2}\thanks{Corresponding author.},
    Bobo Li\textsuperscript{\rm 2},
    Haoyu Zhao\textsuperscript{\rm 3},
    Hao Fei\textsuperscript{\rm 2},
    Mong-Li Lee\textsuperscript{\rm 2},
    Wynne Hsu\textsuperscript{\rm 2}
}
\affiliations{
    \textsuperscript{\rm 1} School of Management Science and Engineering, Anhui University of Finance and Economics, Bengbu, China, \\
    \textsuperscript{\rm 2} National University of Singapore,\\
    \textsuperscript{\rm 3} Wuhan University\\
     liuwei628@aufe.edu.cn, swu@u.nus.edu, libobo@nus.edu.sg
}

\usepackage{bibentry}

\begin{document}

\maketitle

\begin{abstract}
In the AIGC era, generating high-quality 4D content has garnered increasing research attention.  
Unfortunately, current 4D synthesis research is severely constrained by the lack of large-scale 4D datasets, preventing models from adequately learning the critical spatial-temporal features necessary for high-quality 4D generation, thus hindering progress in this domain.  
To combat this, we propose a novel framework that transfers rich spatial priors from existing 3D diffusion models and temporal priors from video diffusion models to enhance 4D synthesis.  
We develop a spatial-temporal-disentangled 4D (STD-4D) Diffusion model, which synthesizes 4D-aware videos through disentangled spatial and temporal latents. 
To facilitate the best feature transfer, we design a novel Orthogonal Spatial-temporal Distributional Transfer (Orster) mechanism, where the spatiotemporal feature distributions are carefully modeled and injected into the STD-4D Diffusion.
Furthermore, during the 4D construction, we devise a spatial-temporal-aware HexPlane (ST-HexPlane) to integrate the transferred spatiotemporal features, thereby improving 4D deformation and 4D Gaussian feature modeling.
Experiments demonstrate that our method significantly outperforms existing approaches, achieving superior spatial-temporal consistency and higher-quality 4D synthesis. 
\end{abstract}

\section{Introduction}

As one of the key fields of computer vision, AIGC has witnessed rapid advancements in the latest decade, evolving from synthesizing static images \cite{rombach2022high,wu2023imagine,wu2024next,wu2025learning} to generating dynamic video content \cite{singer2022make,zuo2024videomv,fei2024dysen,fei2025path}, and achieving comprehensive 3D scene understanding \& generation \cite{chen2024ll3da,shi2023mvdream}. 
Recently, the focus has extended to understanding and generating 4D content \cite{ren2023dreamgaussian4d,zeng2025stag4d,bahmani20244d,liang2024diffusion4d}, representing the next frontier in visual modeling. 
4D synthesis \cite{singer2023text,pmlr-v202-singer23a,wu2024sc4d,miao2025advances,zheng2024unified} has emerged as an important research topic due to its significant potential in practical applications, including animation production \cite{10.1145/2699643}, gaming \cite{li2024dreammesh4d}, and the AR/VR industry \cite{li20244k4dgen}.

\begin{figure}[!t]
    \centering
    \includegraphics[width=\linewidth]{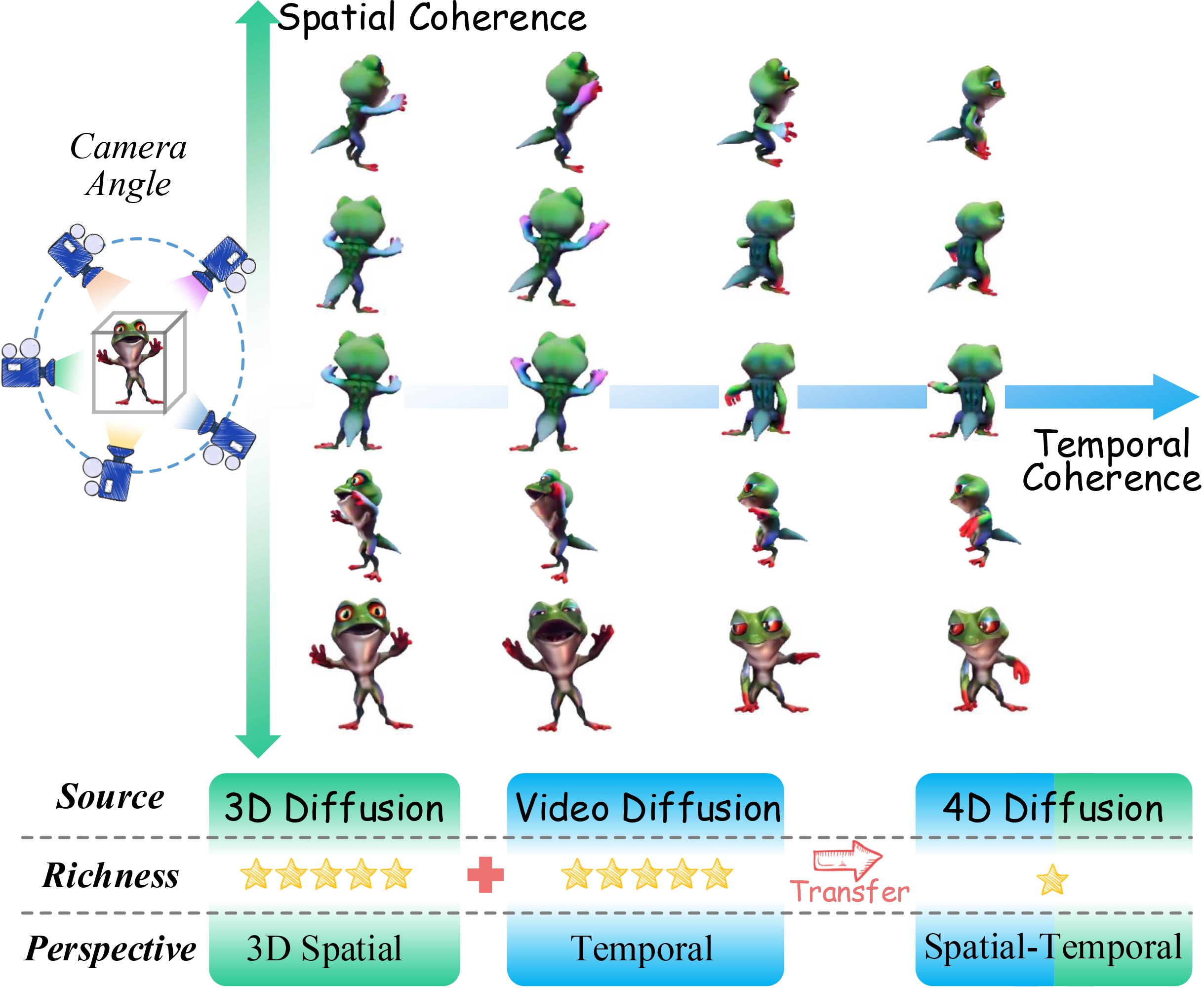}
    \caption{4D generation requires spatial and temporal coherence simultaneously.
    While data resources in 4D are scarce, this paper proposes transferring the 3D spatial prior and temporal prior feature learning from existing resources-rich 3D diffusion and video diffusion, respectively.
    }
    \label{fig:intro}
\end{figure}

Compared to images, videos, and 3D visual content, 4D visual data encompasses the most comprehensive characteristics, integrating both stringent spatial and temporal properties \cite{jiang2023consistent4d}. 
This makes high-quality 4D content generation particularly challenging, as it requires robust spatial-temporal feature modeling \cite{rahamim2024bringing,chen2024ct4d,yuan20244dynamic}. 
Unfortunately, one of the greatest obstacles to progress in this area is the scarcity of labeled 4D datasets \cite{jiang2023consistent4d}, i.e., training a powerful 4D generation model inherently relies on access to large-scale 4D data.
Earlier methods directly train models using the limited available 4D data \cite{deitke2023objaverse}. 
However, the lack of sufficient supervision results in suboptimal spatial-temporal feature modeling, leading to limited 4D generation performance. 
To address this challenge, researchers \cite{liang2024diffusion4d} consider leveraging a pre-trained 3D-aware video diffusion model, where the extensive annotated resources of spatial priors from 3D diffusion models \cite{poole2022dreamfusion} and temporal priors from video diffusion models \cite{zuo2024videomv} are integrated simultaneously.
Fig. \ref{fig:intro} illustrates this intuition.
By performing final fine-tuning on a small amount of 4D data, the model can achieve an overall improvement in 4D generation.

Nevertheless, this research \cite{liang2024diffusion4d} straightforwardly injects the temporal video prior into a 3D backbone diffusion model.  
This method of integrating spatial and temporal priors still faces quite critical issues, which prevent current approaches from fully exploiting the spatial priors from 3D diffusion and the temporal priors from video diffusion.  
On the one hand, directly overlaying temporal features onto 3D spatial features leads to catastrophic forgetting, where the later temporal representation dominates the original spatial features in the backbone 3D diffusion.  
Moreover, this integration method merely transfers feature representations without considering the disentanglement of temporal and spatial features.  
In the 4D generation, the characteristics of time and space follow completely different distributions, which are heterogeneous and orthogonal.  
For example, from the spatial aspect, the distribution describes different parts of a frog's geometry, while at the temporal level, it depicts the motions.  
Conversely, a completely different object with a distinct spatial distribution could perform the same action, thereby sharing the same temporal distribution.  
Thus, during the transfer process, spatial and temporal aspects should be appropriately modeled according to their respective distributions.

\begin{figure*}[!th]
    \centering
    \includegraphics[width=0.99\textwidth]{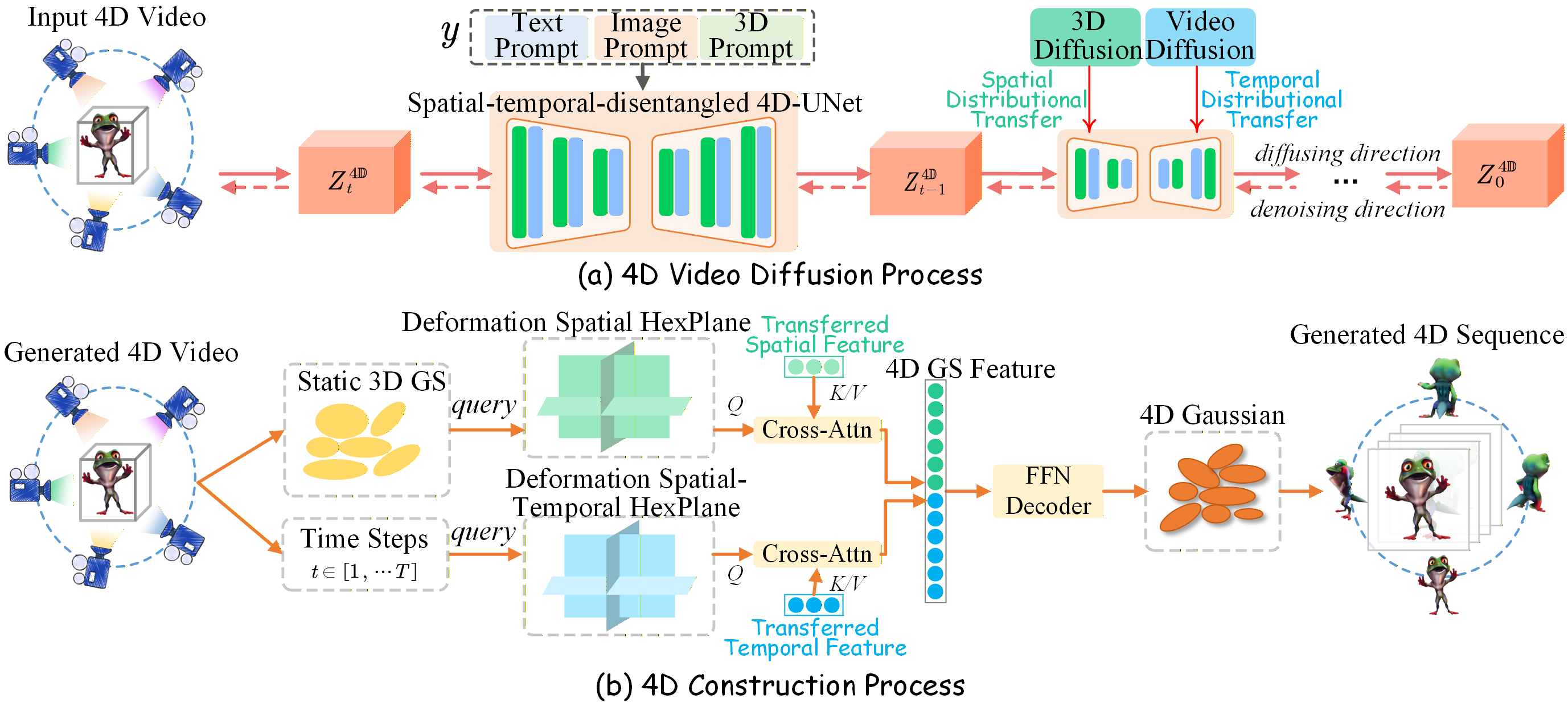}
    \caption{Overall pipeline of our 4D generation framework, including (a) 4D diffusion stage, and (b) 4D construction stage.  
    }
    \label{fig:pipeline}
\end{figure*}

To combat these bottlenecks, this work proposes a novel 4D generation framework that fully leverages the abundant, rich static spatial and dynamic temporal features from 3D and video diffusions, respectively.  
As shown in Fig. \ref{fig:pipeline}, the overall 4D generation pipeline consists of a 4D video diffusion process and a 4D construction process.  
Technically, we first develop a spatial-temporal-disentangled 4D-aware Diffusion (\textbf{STD-4D} Diffusion) model to synthesize 4D-aware videos, wherein a disentangled representation mechanism for spatial and temporal latents is devised.  
The spatial-temporal disentangled 4D-UNet maintains the spatial and temporal feature representations in an interleaved manner, facilitating the separate transfer of spatial and temporal features.  
This is followed by explicit 4D construction using 4D Gaussian Splatting (4DGS) \cite{Wu_2024_CVPR} to generate high-quality 4D assets.  
From the generated 4D video, we decompose \textbf{i)} the spatial hexplane corresponding to the static 3D GS, and \textbf{ii)} the spatiotemporal hexplane corresponding to dynamic sequences, based on which we integrate the transferred spatial and temporal features to form the desired 4DGS feature, and finally decode it to obtain sequences of 4D gaussians, i.e., 4D assets.

Alongside the above system, we design a four-step spatial-temporal-enhanced 4D training process.
Step-\circlenum{1}, the 4D diffusion model is preliminarily pre-trained on limited 4D data to establish a foundational understanding. 
Step-\circlenum{2}, we propose an Orthogonal Spatial-temporal Distributional Transfer (\textbf{Orster}) learning based on a knowledge distillation technique \cite{hinton2015distilling}.
We perform spatial transfer learning and temporal transfer learning simultaneously to inject spatial-temporal knowledge into our spatial-temporal-disentangled 4D-UNet from the host 3D and video diffusions, respectively.
Step-\circlenum{3}, we further perform disentangled spatial-temporal consistency alignment on multi-view video data, ensuring learned spatial and temporal features are fully aligned, thereby guaranteeing the spatial-temporal consistency of the generated 4D content. 
Step-\circlenum{4}, we introduce a phase of conditional 4D generation training, enabling the synthesis of high-quality 4D assets based on various conditions, such as text prompts, images, or static 3D inputs.

To sum up, our main contributions are threefold: \textbf{i)} We present a novel framework to generate high-quality 4D content by transferring spatial-temporal priors from 3D and video diffusion models. \textbf{ii)} We develop a novel spatially-temporally disentangled 4D-aware diffusion model for 4D generation, incorporating an Orthogonal Spatial-temporal Distributional Transfer (Orster) learning mechanism, achieving highly effective knowledge transfer. \textbf{iii)} Extensive qualitative and quantitative experiments demonstrate that our method significantly outperforms existing approaches, generating 4D contents with superior spatial-temporal consistency and rich detail.

\section{Related Work}

\paragraph{Data Scarcity of 4D Generation.}
Recent advances in 3D content generation \cite{xiang2024structured,shi2023mvdream,sun2024dimensionx,li20244k4dgen,xu2025vimorag} have demonstrated remarkable success, driven by techniques such as NeRF \cite{MildenhallSTBRN22}, which revolutionized 3D reconstruction by modeling high-quality static 3D scenes from sparse input views. 
While 3D generation focuses primarily on spatial consistency, 4D generation \cite{ren2024l4gm,xu2024comp4d,wang2024compositional} extends these challenges by incorporating temporal dynamics, making it essential to model both spatial and temporal features simultaneously. 
Unfortunately, one significant challenge in advancing 4D generation is the scarcity of annotated 4D datasets. 
Unlike 3D data, which can often be synthesized or labeled using existing tools, collecting large-scale, high-quality 4D datasets is a labor-intensive process. 
Recent works \cite{pmlr-v202-singer23a,yu20244real} attempt to address this by first learning static 3D shapes and then introducing motion using video diffusion models \cite{WuGPTZBH25,ChuKLHTF25,yang2025not}. 
However, such approaches often result in suboptimal spatiotemporal modeling due to insufficient supervision from limited data. 
In contrast, we leverage well-trained 3D diffusion models and video diffusion models to learn rich spatial priors and temporal priors independently. 
By transferring such knowledge into 4D synthesis, we enable the generation of high-quality 4D content with superior spatial-temporal consistency, effectively overcoming data scarcity.

\paragraph{Spatial-temporal Modeling of 4D Generation.}
Diffusion models \cite{ho2020denoising,croitoru2023diffusion} have emerged as a dominant framework for generative tasks, achieving state-of-the-art (SoTA) performance across multiple domains. 
Notable works in video diffusion include video diffusion models \cite{zhang20244diffusion}, which excel in generating temporally coherent video sequences. 
On the other hand, 3D-aware diffusion models, such as Latent-NeRF \cite{Metzer_2023_CVPR} and 3DiM \cite{watson2023novel}, ensure spatial consistency by generating multi-view-consistent images of static 3D objects. 
While these advancements represent significant progress, adapting these techniques to 4D generation presents unique challenges due to the need for integrating spatial and temporal features across dynamic scenes \cite{0001ZNMSVSTW025}. 
Recent attempts like Diffusion4D \cite{liang2024diffusion4d} tackle this by combining 3D-aware and video diffusion models, but they often struggle with latent entanglement and temporal inconsistencies.
Our work builds upon these advancements by proposing a novel 4D diffusion model that disentangles spatial and temporal latents. 
This allows for dedicated modeling of each aspect while maintaining its coherence in the generation process. 
Through feature distillation, we inject spatial priors from pre-trained 3D diffusion models and temporal priors from pre-trained video diffusion models into our 4D diffusion framework. 
Also, our 4D construction incorporates the SoTA dynamic 4DGS \cite{Wu_2024_CVPR} with HexPlane deformation technology \cite{cao2023hexplane}, enabling the synthesis of high-quality 4D assets from generated videos.

\begin{figure*}[!t]
    \centering
    \includegraphics[width=\textwidth]{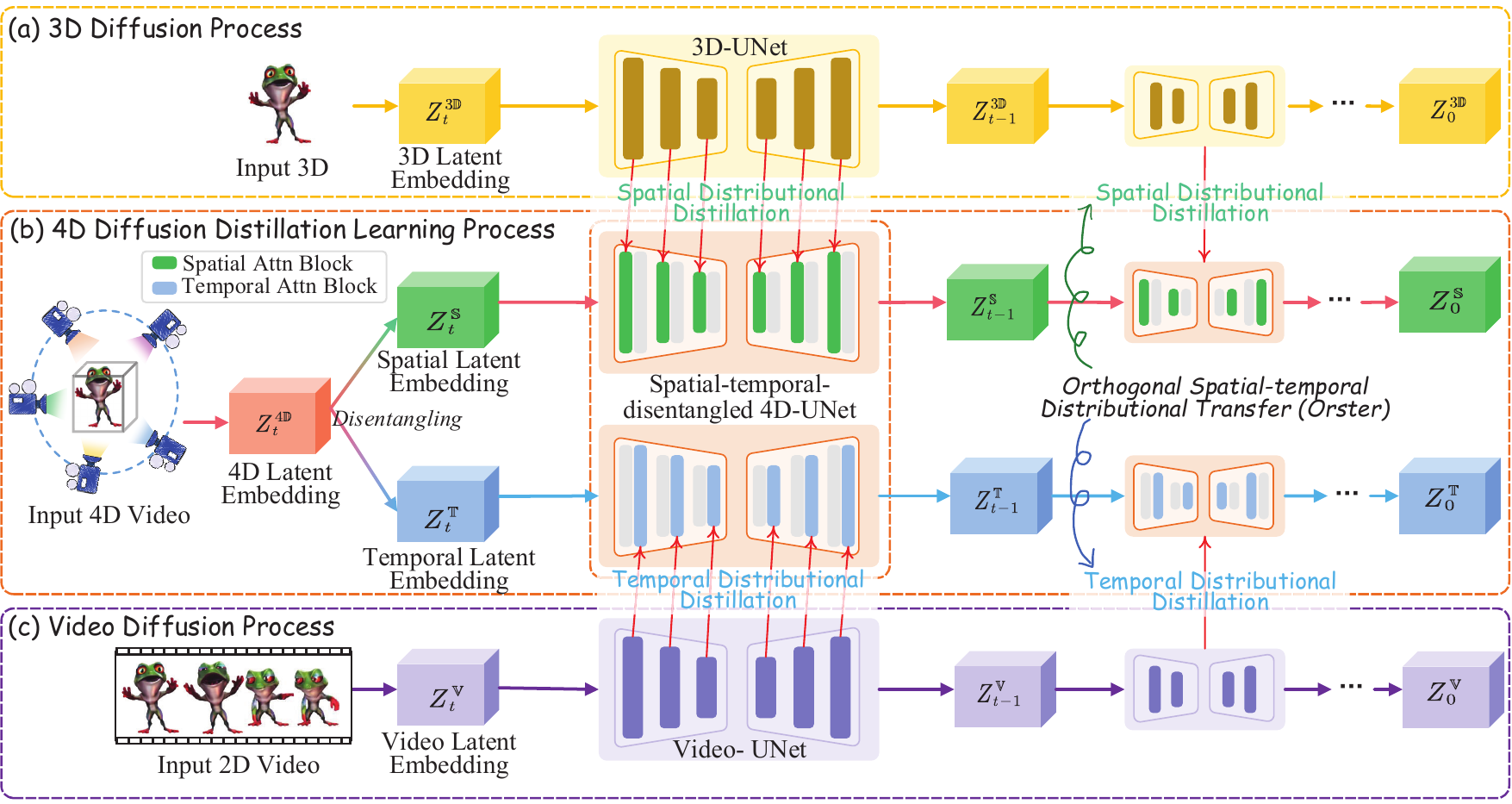}
    \caption{Overview of the knowledge transfer process between external 3D/video Diffusions and our STD-4D Diffusion, where STD-4D Diffusion disentangles the 4D latent into spatial and temporal channels, during which the spatial and temporal features from 3D diffusion and video diffusion are distilled into the spatial and temporal blocks of the 4D-UNet, respectively, via the Orthogonal Spatial-temporal Distributional Transfer (Orster) mechanism (cf. Fig. \ref{fig:Orster}).
    }
    \label{fig:framework}
\end{figure*}

\section{Architecture of 4D Synthesis System}

\paragraph{Task Definition and System Pipeline.}
Our system consists of two components: a 4D Diffusion module and a 4D Construction module.
As seen in Fig. \ref{fig:pipeline}, given a prompt $y$ (can be a text, image or static 3D content), the 4D Diffusion generates an orbital video $\mathcal{V} = \{I_i \in \mathbb{R}^{H \times W \times 3}\}_{i=1}^T$ around a dynamic 3D asset during the denoising process. 
$\mathcal{V}$ consists of $T$ multi-view images, $\mathcal{T} = \{\tau_i\}_{i=1}^T$ along a predefined camera trajectory, where $H$ and $W$ represent the image height and width. 
Then, the 4D Construction generates a high-quality 4D asset $\mathcal{G}^{4D}$ from the orbital video $\mathcal{V}$.

\subsection{Spatial-temporal-disentangled 4D Diffusion}

The core of our 4D-aware diffusion model is a spatial-temporal-disentangled 4D-UNet framework.
Fig. \ref{fig:framework}(b) illustrates this mechanism.
The denoising process begins by encoding the 4D input data into a latent representation using a pre-trained Variational Autoencoder (VAE) \cite{kingma2013auto}. 
Specifically, the VAE first encodes a 4D input $X^{4D}$ into a 4D latents $Z_t^{4D}$.
We then introduce a disentanglement block, another VAE, to disentangle $X^{4D}$ into spatial representation $Z_t^S$ and temporal representation $Z_t^T$:
\begin{align}
Z_t^S &= \text{Disentangle}_{\text{spatial}}(Z_t^{4D}), \\
Z_t^T &= \text{Disentangle}_{\text{temporal}}(Z_t^{4D}).
\end{align}
Each disentangled latent embedding is processed separately through a 4D-UNet, respectively. 
The spatial latent $Z_t^S$ is processed via spatial denoising:
\begin{equation}
Z_{t-1}^S = \epsilon_\theta^S(Z_t^S, t),
\end{equation}
and the $Z_t^T$ is processed via temporal denoising:
\begin{equation}
Z_{t-1}^T = \epsilon_\theta^T(Z_t^T, t),
\end{equation}
where $\epsilon_\theta^S$ and $\epsilon_\theta^T$ are spatial and temporal denoising, respectively.
The disentangled denoised embeddings are then used to update the 4D latent, incorporating the conditional input $y$:
\begin{equation}
Z_{t-1}^{4D} = \text{FFN}(Z_{t-1}^S, Z_{t-1}^T | y).
\end{equation}
Finally, the denoised latent embedding $Z_{t-1}^{4D}$ is decoded back into the 4D domain via the VAE decoder:
\begin{equation}
\hat{X}^{4D} = \text{VAE}_{\text{dec}}(Z_{t-1}^{4D}).
\end{equation}
This disentangled modeling ensures that spatial and temporal dynamics are effectively learned and aligned, during which we can inject the spatial and temporal features into the above process, respectively.
This will be elaborated later.

\begin{figure}[!t]
    \centering
    \includegraphics[width=\linewidth]{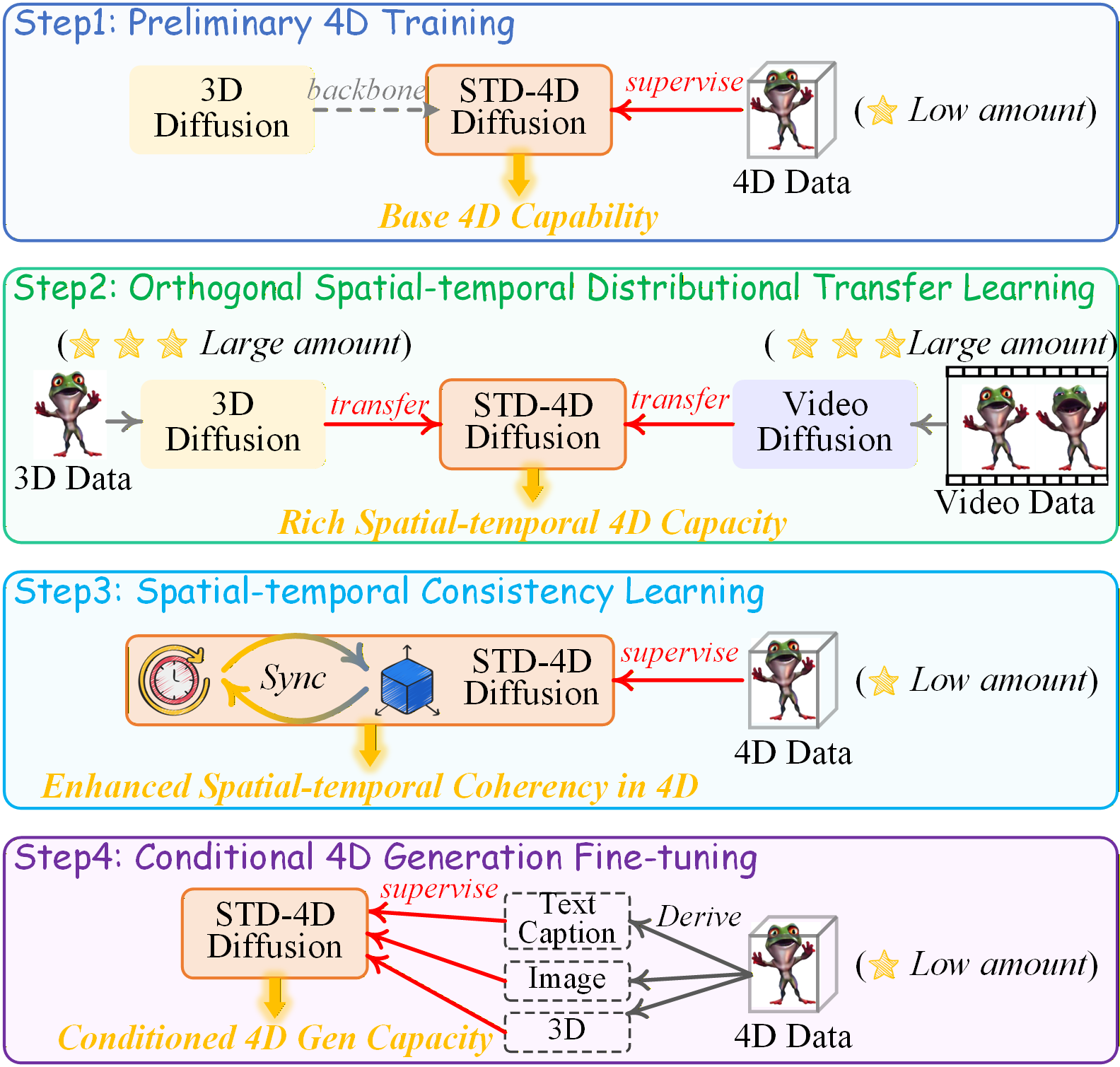}
    \caption{
    Four-stage STD-4D Diffusion training.
    }
    \label{fig:training}
\end{figure}

\subsection{Gaussian Deformation for 4D Construction with Spatial-temporal-aware HexPlane}

The construction module incorporates the SoTA dynamic 4DGS \cite{Wu_2024_CVPR} for 4D synthesis.
After generating an orbital video $\mathcal{V}$, we extract spatial and temporal anchors, represented by static 3D Gaussians and their associated temporal sequence. 
To model the dynamic 4D representation, we employ a HexPlane \cite{cao2023hexplane} structure that encodes 4D spatial-temporal information by decomposing the 4D field into six deformation feature planes, spanning each pair of coordinate axes.
Given a query point $(x, y, z, t)$, the HexPlane predicts the transformation parameters, position displacement $\Delta \mathbf{p}$, rotation $\mathbf{R}$, and scale $\mathbf{s}$, for each Gaussian anchor $\mathcal{G}$:
\setlength{\abovedisplayshortskip}{0pt}
\setlength{\belowdisplayshortskip}{0pt}
\begin{equation}
\Delta \mathbf{p}, \mathbf{R}, \mathbf{s} \gets \text{HexPlane}( D(x, y, z, t) ),
\end{equation}
where $D(\cdot)$ represents the transformations within HexPlane, and the process dynamically adjusts the Gaussians to capture object motion across time.
To obtain these parameters more accurately, we consider a spatial-temporal-aware HexPlane (ST-HexPlane), which utilizes the transferred spatial prior \(\mathbf{O}^{s}\) and temporal prior \(\mathbf{O}^{t}\) from the existing 3D diffusion and video diffusion.
\begin{align}
\label{feat-prior} \mathbf{V}^{s/t} = \text{Attn}(Q,K,V)& = \text{Attn}( D(x, y, z, t) , \mathbf{O}^{s/t}, \mathbf{O}^{s/t}) \,, \\
\Delta \hat{\mathbf{p}}, \hat{\mathbf{R}}, \hat{\mathbf{s}} \gets& \text{ST-HexPlane}([\mathbf{V}^{s} ; \mathbf{V}^{t}]),
\end{align}
The updated Gaussians are then decoded for the 4D asset:
\setlength{\abovedisplayshortskip}{0pt}
\setlength{\belowdisplayshortskip}{0pt}
\begin{equation}
\mathcal{G}^{4D} = \text{Decoder}(\mathcal{G} + \Delta \mathbf{p}, \mathbf{R}, \mathbf{s}).
\end{equation}
Refer to \cite{ren2023dreamgaussian4d} for more decoding technical details.
The process, as illustrated in Fig. \ref{fig:pipeline}(b), enables precise modeling of 4D dynamics by leveraging HexPlane's efficient spatial-temporal factorization and 4DGS's ability to represent high-fidelity dynamic structures.

\section{Spatial-temporal-enhanced 4D Learning}

\paragraph{Overview.}
Building upon the above 4D synthesis system, particularly with the Spatially-temporally Disentangled 4D diffusion, in this section, we propose spatial-temporal aware learning.  
Our key idea is to transfer the rich spatial and temporal features learned by 3D Diffusions and Video Diffusions, which benefit from extensive training signals, into our 4D diffusion framework (specifically the 4D U-Net), compensating for the lack of sufficient spatial-temporal signals required for 4D Diffusion generation.  
Technically, we design four training stages: 1) preliminary 4D training, 2) orthogonal spatial-temporal distributional transferring, 3) spatial-temporal consistency training, and 4) conditional 4D generation training.
Fig. \ref{fig:training} illustrates the overall stages.

\subsection{Preliminary 4D Training}
As the first step, before performing transfer learning, we enable the backbone diffusion model with preliminary dynamic 4D generation capabilities.  
Considering that 3D and 4D are closely related modalities, we adopt a pre-trained 3D-aware video diffusion model \cite{zuo2024videomv} as our backbone. 
Then, we train it on Objaverse \cite{deitke2023objaverse}, consisting of multi-view dynamic orbital image sequences.  
This object is marked as $\mathcal{L}_{\text{ldm}}=||\epsilon_t - \epsilon(Z_t,Y,t)||$, where $Z_t$ is the 4D latent.
Due to the very limited data size, the 4D features learned in this stage are quite restricted, serving primarily as foundational training to establish a baseline for subsequent transfer learning.

\subsection{Orthogonal Spatial-temporal Distributional Transfer (Orster) Learning}

Based on the distillation learning \cite{hinton2015distilling}, we now perform Orster learning to gain rich spatiotemporal features.
As shown in Fig. \ref{fig:framework}, building on our STD-4D Diffusion framework, we distill the spatial and temporal features from external well-trained 3D diffusion and video diffusion models, respectively.
Via our Orster mechanism, the spatial geometry features from the UNet of the 3D diffusion model are injected into the spatial blocks of the 4D-UNet, while the temporal features from the UNet of the video diffusion model are fused into the temporal blocks of the 4D-UNet.

\begin{figure}[!t]
    \centering
    \includegraphics[width=\linewidth]{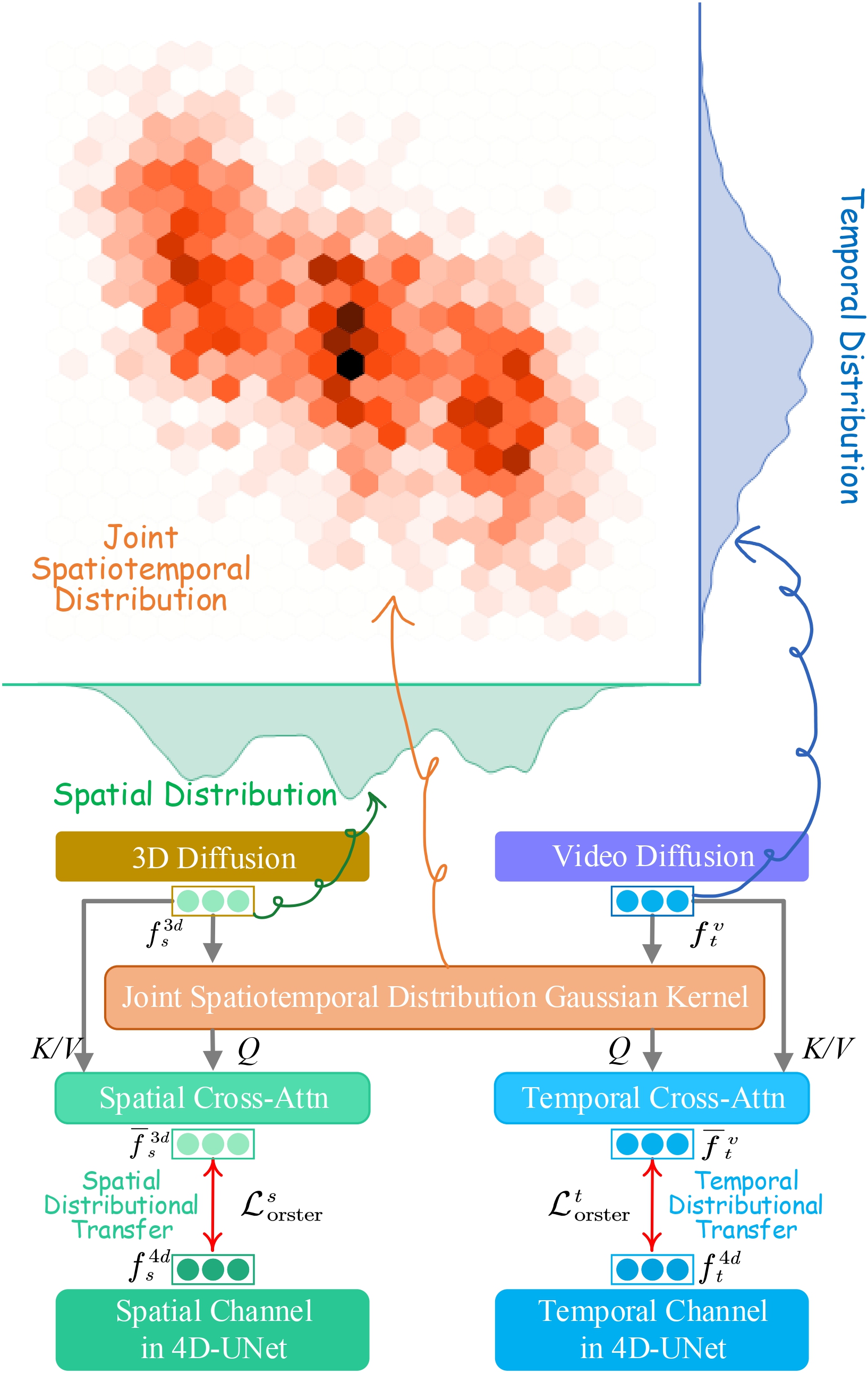}
    \caption{
    A closer look at the Orthogonal Spatial-temporal Distributional Transfer (Orster) learning module.
    }
    \label{fig:Orster}
\end{figure}

As mentioned earlier, the spatial and temporal distributions should be modeled separately during the distillation process. 
However, within the same 4D scene, time and space also exhibit a reasonable joint distribution. 
To capture the intricate interaction between spatial and temporal feature distributions, we propose Orster as shown in Fig. \ref{fig:Orster}.
First, for any spatial embedding $f_s$ from host 3D Diffusion and temporal embedding $f_t$ from host video Diffusion, we define the joint spatiotemporal distribution Gaussian Kernel as:
\setlength\abovedisplayskip{3pt}
\setlength\belowdisplayskip{3pt}
\begin{equation}\label{kernal}
\begin{split}
\kappa(f_s, f_t) = \exp\Biggl( \frac{-1}{2} \Biggl(
\frac{\|f_s - g_s\|^2}{\sigma_s^2} + \frac{\|f_t - g_t\|^2}{\sigma_t^2} \\
+ \frac{2\alpha\,\langle f_s - g_s,\, f_t - g_t \rangle}{\sigma_{st}^2}
\Biggr) \Biggr) \,,
\end{split}
\end{equation}
where \((g_s, g_t)\) are the mean spatial and temporal features of all \((f_s, f_t)\), scale parameters $\sigma_s, \sigma_t, \sigma_{st}$ and $\alpha$ are learnable.
Then, we compute the resulting feature representation via Spatial/Temporal Cross-Attention, respectively:
\setlength\abovedisplayskip{3pt}
\setlength\belowdisplayskip{3pt}
\begin{align}
\label{spt-feat}  \bar{f_s^{3d}} = & \text{Spt-Attn}({f_s^{3d}}, \kappa, f_s^{3d} ) \,, \\
\label{tmp-feat}  \bar{f_t^{v}} = & \text{Tmpr-Attn}({f_t^{v}}, \kappa, f_t^{v} ) \,.
\end{align}
Lastly, we conduct the distillation over $\bar{f}$ and ${f}$:
\setlength\abovedisplayskip{3pt}
\setlength\belowdisplayskip{3pt}
\begin{equation}\label{orster-loss}
\begin{aligned} \mathcal{L}_{\text{orster}}=\underbrace{\lambda_o ||f^{4d}_s - \bar{f_s^{3d}} ||}_{\textcolor{BlueGreen}{\mathcal{L}^s_{\text{orster}}}}
    +
    \underbrace{(1-\lambda_o) ||f^{4d}_t - \bar{f_t^{v}} ||}_{\textcolor{NavyBlue}{\mathcal{L}^t_{\text{orster}}}},
\end{aligned}
\end{equation}
where $\lambda_o$ is a term weight.
This approach helps the effective transfer of spatial-temporal features into the 4D diffusion model. 
Once the transfer learning is completed, we merge the spatial and temporal blocks from the two channels into a unified 4D-UNet in a complete and integrated view.

\subsection{Spatial-temporal Consistency Learning}

Despite obtaining sufficiently high-quality spatiotemporal features in the previous step via sophisticated modeling, inconsistencies may still arise because the spatial and temporal features are extracted from different sources. 
To address this, we perform spatial-temporal consistency training to further refine and align the learned features. 
We leverage the same multi-view 4D video dataset to jointly tune the spatial and temporal alignment, ensuring coherent integration across both dimensions. 
Our overall objective is given by:
\setlength\abovedisplayskip{2pt}
\setlength\belowdisplayskip{2pt}
\begin{equation}
\mathcal{L}_{\text{const}} = \lambda_{\text{rec}} \mathcal{L}_{\text{rec}} + \lambda_{\text{perc}} \mathcal{L}_{\text{perc}} + \lambda_{\text{temp}} \mathcal{L}_{\text{temp}} + \lambda_{\text{align}} \mathcal{L}_{\text{align}} \,,
\end{equation}
where the reconstruction loss \(\mathcal{L}_{\text{rec}} = \|I_{\text{pred}} - I_{\text{gt}}\|_1\) ensures accurate view-wise reconstruction, 
and the perceptual loss \(\mathcal{L}_{\text{perc}} = \sum_l \|\phi_l(I_{\text{pred}}) - \phi_l(I_{\text{gt}})\|_2^2\) (with \(\phi_l(\cdot)\) denoting features from a pretrained network) captures high-level visual fidelity. 
Also, the temporal smoothness loss \(\mathcal{L}_{\text{temp}} = \|F_t^{(t+1)} - F_t^{(t)}\|_2^2\) promotes consistency across consecutive frames, 
and the feature alignment loss \(\mathcal{L}_{\text{align}} = 1 - \frac{\langle F_s, F_t \rangle}{\|F_s\| \cdot \|F_t\|}\) enforces coherent integration between spatial and temporal features. 
Iterative optimization of \(\mathcal{L}_{\text{const}}\) gradually refines the spatiotemporal representations, ensuring precise alignment and overall coherence in the generated 4D content.

\subsection{Conditional 4D Generation Fine-tuning}
Our framework supports various prompts \(y\) as generation conditions, including texts, images, or static 3D content.  
For text conditions, text embeddings \(T\) are extracted using the CLIP model and injected into the 4D-UNet via a cross-attention mechanism.  
For image conditions, the first-view image \(I_0\) of the orbital video \(V\), captured at timestamp 0, is used as the reference image, which is injected into the 4D U-Net through the cross-attention mechanism.  
For static 3D conditions, the video \(V\) is used as the reference, where video features are extracted by a pre-trained encoder and fed into the diffusion model.  
Thus, we perform diverse Conditional 4D Generation Training, enabling the model to learn to synthesize high-quality 4D assets from various conditions.
Specifically, we optimize the following objective:
\setlength\abovedisplayskip{3pt}
\setlength\belowdisplayskip{3pt}
\begin{equation}
\mathcal{L}_{\text{cond}} = \mathbb{E}_{Z_0, (T|I_0|V), t, \epsilon}\left[\|\epsilon - \epsilon_\theta(Z_t, (T|I_0|V), t)\|^2\right] \,.
\end{equation}

\subsection{4D Construction Optimization}
In addition to training our STD-4D Diffusion module, the 4D Construction process (cf. Fig.~\ref{fig:pipeline}(b)) also requires optimization. 
First, for the first-view image \(I_0\) in the video \(V\), we use a pretrained 3D-aware video Diffusion model to generate an orbital-view video \(\bar{V}'\) of a static 3D object. 
As the 4D-aware video Diffusion model is finetuned from that pretrained model, there is a high 3D geometric consistency between \(\bar{V}'\) and \(V\). 
Hence, we train with this data. 
For the ST-HexPlane's Deformation Field optimization, we follow~\cite{ren2023dreamgaussian4d}, fix the camera to the reference view, and minimize the Mean Squared Error (MSE) between the rendered image and the driving video frame at each timestamp,
\setlength\abovedisplayskip{3pt}
\setlength\belowdisplayskip{3pt}
\begin{equation}
\mathcal{L}_{\text{hex}} = \frac{1}{\tau} \sum_{t=1}^{\tau} \| f(\phi(S,\tau),\, o_{\text{ref}}) - I_{\text{ref}}^\tau \|^2.
\end{equation}
We then refine the 4DGS using only \(V\) to enhance spatiotemporal awareness; thanks to the 4D consistency in our generated videos, precise pixel-level matching across different views and timestamps can be achieved. 
Following~\cite{liang2024diffusion4d}, we adopt \(L_1\) and \(L_{\text{lpips}}\)~\cite{zhang2018unreasonable} losses for optimization and introduce a depth smoothness loss~\cite{niemeyer2022regnerf} as a regularizer to reinforce geometric smoothness. 
The total loss is formulated as:
\setlength\abovedisplayskip{3pt}
\setlength\belowdisplayskip{3pt}
\begin{equation}
\mathcal{L}_{\text{gs}} = \lambda_{l1}\,\mathcal{L}_{l1} \;+\; \lambda_{\text{lpips}}\,\mathcal{L}_{\text{lpips}} \;+\; \lambda_{\text{dep}}\,\mathcal{L}_{\text{dep}} + \lambda_{\text{hex}}\,\mathcal{L}_{\text{hex}} \,.
\end{equation}

\section{Experiment}

\subsection{Settings}

\textbf{Implementation Details.} 
We adopt VideoMV \cite{zuo2024videomv} as the backbone 3D diffusion model, and ModelScopeT2V \cite{wang2023modelscope}, and I2VGen-XL \cite{zhang2023i2vgenxl} as text-to-video and image-to-video diffusion backbone, respectively.
Our 4D-aware video diffusion model is trained for 5,000 iterations. 
During the sampling phase, we condition the model on text prompts, front-view images, or orbital videos of static 3D assets starting from the front view.  
In the 4D construction stage, we optimize the 4DGS representation in two phases: a coarse optimization phase consisting of 6,000 iterations and a fine optimization phase consisting of 3,000 iterations. 
We employ the Consistent4D \cite{jiang2023consistent4d} as a test set.

\noindent\textbf{Evaluations.} 
We consider both automatic quantitative evaluation and human qualitative evaluation.  
Following \cite{liang2024diffusion4d}, we adopt CLIP-O, where 36 uniformly rendered orbital views of the generated 4D assets are used as targets, and CLIP-F, which evaluates scores using only front-view images as targets. 
Also, we utilize LPIPS, PSNR, SSIM, and FVD to assess the appearance, texture quality, and the spatial-temporal consistency of the generated visuals.  
The baselines include 4DFY \cite{bahmani20244d}, Animate124 \cite{zhao2023animate124}, Diffusion4D \cite{liang2024diffusion4d}, 4DGen \cite{yin20234dgen}, and STAG4D \cite{zeng2025stag4d}.

\begin{table}[!t]
\fontsize{9}{11}\selectfont 
\setlength{\tabcolsep}{0.3mm}
\centering
\begin{tabular}{lcccccc}
\hline
\textbf{Model} & \textbf{CLIP-F$\uparrow$} & \textbf{CLIP-O$\uparrow$} & \textbf{SSIM$\uparrow$} & \textbf{PSNR$\uparrow$} & \textbf{LPIPS$\downarrow$} & \textbf{FVD$\downarrow$} \\
\hline
\multicolumn{7}{l}{$\bullet$ \textbf{Text-to-4D}} \\
4DFY & 0.78 & 0.61 & - & 14.2 & 0.23 & 1042.3 \\
Animate124 & 0.75 & 0.58 & - & 15.0 & 0.21 & 720.5 \\
Diffusion4D & 0.82 & 0.69 & - & 15.3 & 0.20 & 684.0 \\
\rowcolor{uclablue} \bf Ours & \bf 0.85 & \bf 0.72 & - & \bf 16.8 & \bf 0.19 & \bf 523.4 \\
\hline
\multicolumn{7}{l}{$\bullet$ \textbf{Image-to-4D}} \\
4DGen & 0.84 & 0.71 & 0.69 & 14.4 & 0.31 & 736.6 \\
STAG4D& 0.86 & 0.72 & 0.76 & 15.2 & 0.27 & 675.4 \\
Diffusion4D & 0.90 & 0.80 & 0.82 & 16.8 & 0.19 & 490.2 \\
\rowcolor{uclablue} \bf Ours & \bf 0.93 & \bf 0.82 & \bf 0.84 & \bf 17.3 & \bf 0.17 & \bf 477.7 \\
\hline
\multicolumn{7}{l}{$\bullet$ \textbf{3D-to-4D}} \\
Diffusion4D& 0.91 & 0.81 & 0.83 & 17.2 & 0.18 & 482.4 \\
\rowcolor{uclablue} \bf Ours & \bf 0.95 & \bf 0.84 & \bf 0.85 & \bf 17.6 & \bf 0.16 & \bf 465.3 \\
\hline
\end{tabular}
\caption{Main results under different task settings.}
\label{main-res}
\end{table}

\begin{table}[!t]
\fontsize{9}{11}\selectfont 
\setlength{\tabcolsep}{0.0mm}
\centering
\begin{tabular}{lccccc}
\hline
\textbf{} & \textbf{CLIP-F$\uparrow$} & \textbf{CLIP-O$\uparrow$} & \textbf{PSNR$\uparrow$} & \textbf{LPIPS$\downarrow$} & \textbf{FVD$\downarrow$} \\
\hline
\textbf{Ours (complete)} & 0.93 & 0.82 & 17.3 & 0.17 & 477.7 \\
\hline
\multicolumn{6}{l}{$\bullet$ \textbf{4D Diffusion}} \\
w/o Disentangling & 0.77 & 0.65 & 15.0 & 0.31 & 596.7 \\
\quad w/o Spt Channel & 0.87 & 0.77 & 16.3 & 0.25 & 501.1 \\
\quad w/o Tmp Channel & 0.86 & 0.78 & 16.5 & 0.24 & 549.3 \\
\cdashline{1-6}
\multicolumn{6}{l}{$\bullet$ \textbf{4D Construction}} \\
w/o Spt Feat. (Eq. \ref{feat-prior}) & 0.86 & 0.76 & 16.6 & 0.24 & 522.4  \\
w/o Tmp Feat. (Eq. \ref{feat-prior}) & 0.89 & 0.81 & 16.9 & 0.21 & 578.5  \\
\hline
\end{tabular}
\caption{Ablation study on system architecture.}
\label{tab:ablation1}
\end{table}

\begin{figure*}[!t]
    \centering
    \includegraphics[width=\textwidth]{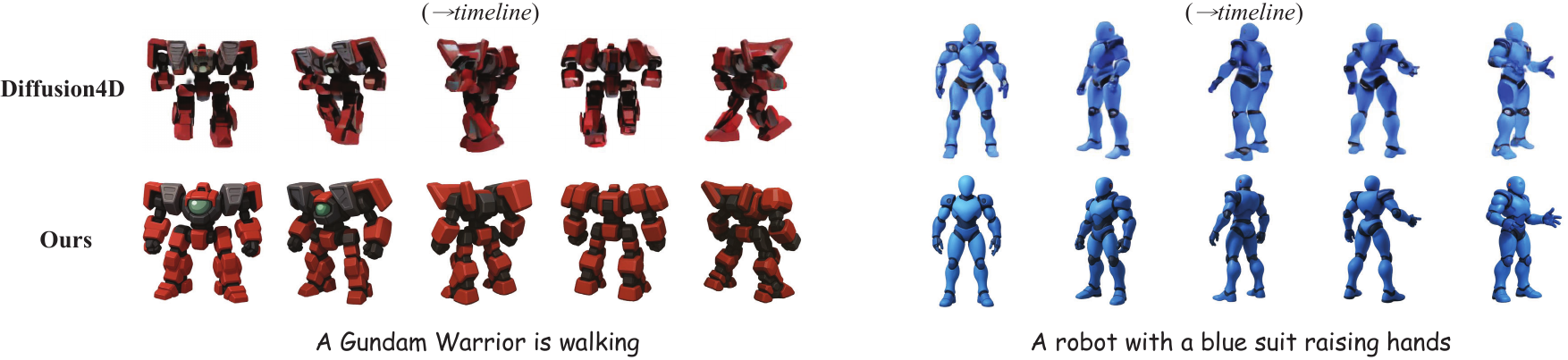}
    \caption{4D generation comparisons with SoTA baseline. Best viewing via zooming in.
}
    \label{fig:compr-baseline}
\end{figure*}

\subsection{Overall Results and Ablation Study}

\paragraph{Overall Results.} 
We compare our approach with several SoTA baselines under text-to-4D, image-to-4D, and 3D-to-4D settings.  
Table \ref{main-res} presents the comprehensive results, from which we can derive the following key observations.  
First, we observe that the overall performance varies across different settings, with the general trend being 3D condition $>$ Image $>$ Text. 
This is because the former provides richer priors and initial information for feature generation.  
Second, our proposed system consistently outperforms all strong-performing baselines across all settings and metrics, directly demonstrating its effectiveness comprehensively.

\paragraph{Ablation Study on System Architecture.} 
In Table \ref{tab:ablation1}, we present the system ablation to explore the fine-grained contributions of different modules.  
First, for the backbone 4D Diffusion, we analyze the disentangling mechanism, the roles of the spatial block, and the temporal block. 
When either of these blocks is removed, we observe varying degrees of performance degradation, highlighting their essential impacts, where the overall spatiotemporal disentangling mechanism shows the biggest influence.  
Also, for the 4D Construction module, we see that both the spatial feature and temporal feature exhibit varied yet non-negligible impacts.

\paragraph{Ablation Study on Learning Strategy.} 
Next, we analyze the impact of training strategies, cf. Table \ref{tab:ablation2}. 
Removing the preliminary training ($\mathcal{L}_{\text{ldm}}$), or spatiotemporal consistency learning ($\mathcal{L}_{\text{const}}$) and 4D construction training ($\mathcal{L}_{\text{gs}}$) result in varied performance drop.
Notably, the proposed Orster learning ($\mathcal{L}_{\text{orster}}$) contributes most significantly.
Stepping into Orster, we see that the joint spatiotemporal distribution Gaussian kernel and also the spatial \& temporal attention mechanisms play an important role.

\begin{table}[!t]
\fontsize{9}{11}\selectfont 
\setlength{\tabcolsep}{0.0mm}
\centering
\begin{tabular}{lccccc}
\hline
\textbf{} & \textbf{CLIP-F$\uparrow$} & \textbf{CLIP-O$\uparrow$} & \textbf{PSNR$\uparrow$} & \textbf{LPIPS$\downarrow$} & \textbf{FVD$\downarrow$} \\
\hline
\textbf{Ours (complete)} & 0.93 & 0.82 & 17.3 & 0.17 & 477.7 \\
\hline
w/o Pre. 4D ($\mathcal{L}_{\text{ldm}}$) & 0.82 & 0.60 & 15.7 & 0.38 & 601.6 \\
\cdashline{1-6}
 w/o Orster ($\mathcal{L}_{\text{orster}}$)  & 0.40 & 0.32 & 12.7 & 0.36 & 668.3 \\
\quad  w/o Spt ($\lambda_o$=0) & 0.67 & 0.50 & 15.0 & 0.40 & 503.9 \\
\quad  w/o Tmp ($\lambda_o$=1) & 0.59 & 0.55 & 14.3 & 0.34 & 557.7 \\
w/o ST Kernal (Eq. \ref{kernal}) & 0.84 & 0.67 & 16.3 & 0.44 & 566.4  \\
w/o Spt Attn. (Eq. \ref{spt-feat}) & 0.80 & 0.70 & 15.7 & 0.40 & 536.8  \\
w/o Tmp Attn. (Eq. \ref{tmp-feat}) & 0.78 & 0.71 & 16.0 & 0.41 & 560.3  \\
\cdashline{1-6}
w/o ST Consis. ($\mathcal{L}_{\text{const}}$) & 0.85 & 0.72 & 16.1 & 0.23 & 542.2 \\
\cdashline{1-6}
w/o 4D Constr. ($\mathcal{L}_{\text{gs}}$) & 0.68 & 0.60 & 15.3 & 0.37 & 587.4 \\
\hline
\end{tabular}
\label{tab:ablation2}
\caption{Ablation study on learning strategies.}
\end{table}

\subsection{Qualitative Results with Visualizations}

We provide visual comparisons to better aid the understanding of how advanced our system can be in 4D generation.
As shown in Fig.~\ref{fig:compr-baseline}, our system generates accurate and very realistic 4D assets with much more excellent spatial and temporal consistency. 
While the baseline Diffusion4D model produces 4D assets, their results often exhibit quite low-quality geometry of appearances or nearly imperceptible motion. 
In contrast, our approach delivers high-fidelity and dynamic actions, and very fine details and more refined appearances

\section{Conclusion}

This paper proposes a novel framework for high-quality 4D generation by transferring rich spatial priors from 3D diffusion models and temporal priors from video diffusion models.
We develop a Spatial-Temporal-Disentangled 4D Diffusion model that synthesizes 4D-aware videos through disentangled spatial and temporal latent spaces.
To facilitate effective cross-modal prior transfer, we introduce the Orster mechanism, which injects spatiotemporal feature distributions into the STD-4D Diffusion.
Extensive experiments demonstrate that our approach achieves superior spatial–temporal consistency and overall higher 4D generation quality than existing baseline methods.

\section*{Acknowledgments}
This work was supported by the National Natural Science Foundation of China (Grant No. 62202001), and the Ministry of Education, Singapore, under its MOE AcRF TIER 3 Grant (MOE-MOET32022-0001).

\bigskip

\bibliography{aaai2026}

\end{document}